%% file: main.tex
\documentclass[11pt]{article}

\usepackage{acl}
\usepackage{times}
\usepackage{latexsym}
\usepackage[T1]{fontenc}
\usepackage[utf8]{inputenc}
\usepackage{microtype}
\usepackage{booktabs}
\usepackage{amsfonts}
\usepackage{amsmath}
\usepackage{nicefrac}
\usepackage{xcolor, colortbl}
\usepackage{graphicx}
\usepackage{subcaption}
\usepackage{enumitem}
\usepackage{tikz}
\usetikzlibrary{positioning, arrows.meta, backgrounds, fit, calc}
\usepackage{placeins}
\usepackage{tcolorbox}

\title{When Valid Signals Fail: \\Regime Boundaries Between LLM Features and RL Trading Policies}

\author{%
  Zhengzhe Yang \\
  Independent Researcher \\
  \texttt{zhengzhe.yang@outlook.com} \\
}

\begin{document}

\maketitle

\begin{abstract}
Can large language models (LLMs) generate continuous numerical features that improve reinforcement learning (RL) trading agents? We build a modular pipeline where a frozen LLM serves as a stateless feature extractor, transforming unstructured daily news and filings into a fixed-dimensional vector consumed by a downstream PPO agent. We introduce an automated prompt-optimization loop that treats the extraction prompt as a discrete hyperparameter and tunes it directly against the Information Coefficient---the Spearman rank correlation between predicted and realized returns---rather than NLP losses. The optimized prompt discovers genuinely predictive features (IC above~0.15 on held-out data). However, these valid intermediate representations do not automatically translate into downstream task performance: during a distribution shift caused by a macroeconomic shock, LLM-derived features add noise, and the augmented agent under-performs a price-only baseline. In a calmer test regime the agent recovers, yet macroeconomic state variables remain the most robust driver of policy improvement. Our findings highlight a gap between feature-level validity and policy-level robustness that parallels known challenges in transfer learning under distribution shift.
\end{abstract}

\section{Introduction}
\label{sec:intro}

Recent work increasingly applies large language models to financial decision-making, whether as end-to-end trading agents or as modular sentiment classifiers feeding downstream models. While these modular pipelines separate the language model from the trading algorithm, they frequently suffer from an \emph{objective mismatch}: the LLM is optimized against standard NLP losses (e.g., classification accuracy or sentiment polarity) rather than downstream financial utility. Consequently, it remains difficult to guarantee that the extracted narratives form a robust state representation for a continuous trading policy.

We address this gap by maintaining strict architectural separation while directly aligning the feature extraction process with a financial objective. The frozen LLM acts as a \emph{stateless feature extractor}: given a bundle of news articles and SEC filings for a ticker on day~$d$, it emits a fixed-length numerical vector (sentiment, impact, conflict flags, etc.). A separate PPO agent then consumes this vector alongside price data and macroeconomic indicators to make portfolio decisions. This design ensures that the intermediate representations are genuinely predictive, allowing us to evaluate the LLM's true contribution in isolation.

\begin{figure*}[t]
\centering
\includegraphics[width=\textwidth]{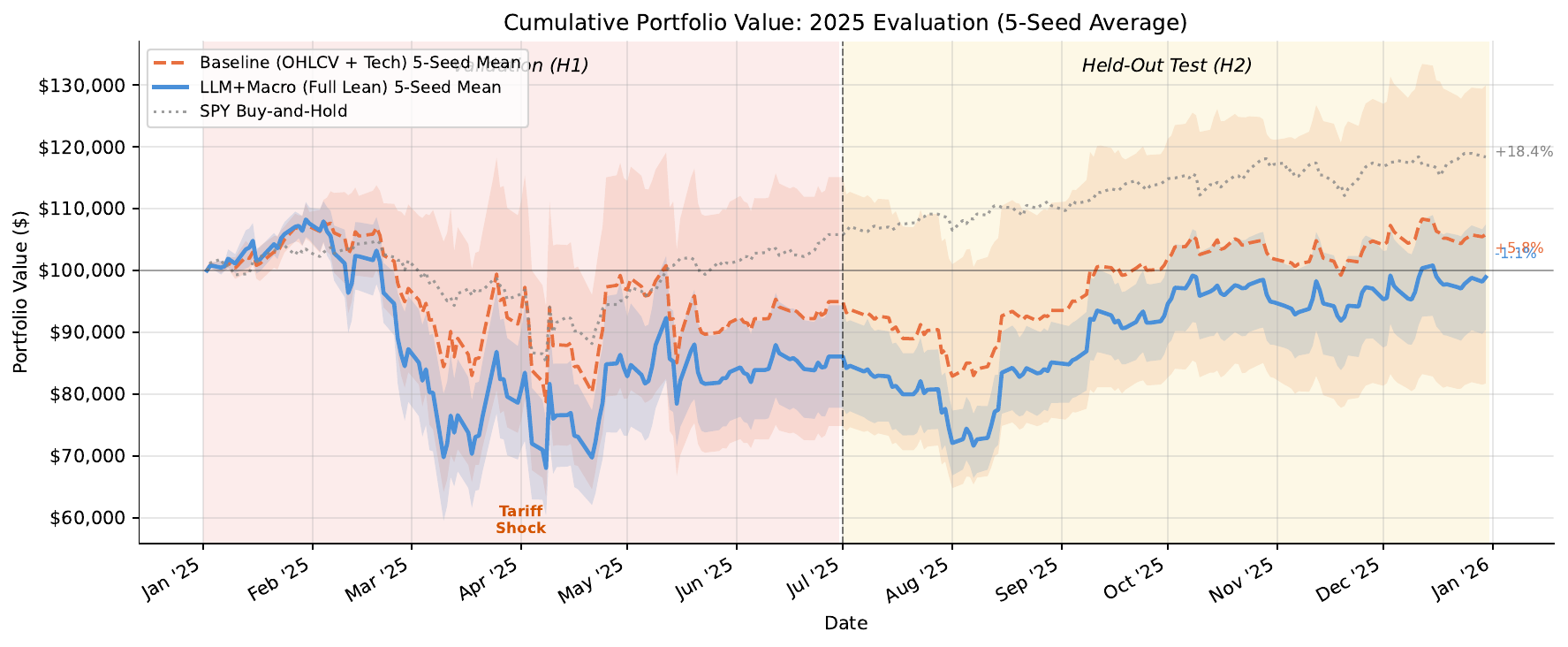}
\caption{Cumulative portfolio value across 2025.  Solid lines: 5-seed mean; shaded: $\pm 1$ std.  H1 (red) captures a tariff-driven shock; H2 (yellow) is calmer.  The regime split isolates when LLM features succeed and fail.}
\label{fig:portfolio_hero}
\end{figure*}

Our contributions are:
\begin{enumerate}[nosep]
  \item \textbf{Prompt-as-hyperparameter optimization.}  We introduce a mutation--evaluation--selection loop that treats the LLM extraction prompt as a discrete hyperparameter and optimizes it against the Information Coefficient (IC)---the rank correlation between predicted and realized returns---rather than NLP losses like BLEU or accuracy.  The winning prompt improves IC from $-0.024$ to $+0.104$ (Table~\ref{tab:prompt_mutations}).
  \item \textbf{Feature-validity-to-policy-utility gap.}  We show that valid intermediate LLM representations do not automatically translate into downstream RL performance.  The gap is regime-dependent: under distribution shift caused by a macroeconomic shock, news-derived features add noise rather than signal.
  \item \textbf{Multi-regime ablation on held-out data.}  A controlled four-configuration ablation (Baseline, LLM-only, Macro-only, LLM+Macro) across a volatile validation period (H1~2025) and a calmer held-out test period (H2~2025) reveals that macroeconomic state variables are the most reliable driver of policy robustness (Figure~\ref{fig:portfolio_hero}).
\end{enumerate}

\section{Related Work}
\label{sec:related}

\paragraph{Reinforcement Learning for Trading.}
Deep RL has been applied to portfolio management and order execution with increasing sophistication~\citep{hambly2023recent}. FinRL~\citep{liu2020finrl} provides a standardized library for training PPO, A2C, and DDPG agents on market environments.  A persistent challenge is non-stationarity: the data distribution shifts between training and deployment, a problem well-studied in the broader RL literature as distributional shift~\citep{kumar2020conservative}.  In finance, these shifts are driven by macroeconomic regime changes~\citep{ang2002international}, and recent work has explored conditioning RL policies on detected regimes~\citep{sun2023market}.

\paragraph{LLMs in Finance.}
BloombergGPT~\citep{wu2023bloomberggpt} demonstrated that domain-specific pre-training improves financial NLP benchmarks.  FinGPT~\citep{yang2023fingpt} pursued the same goal with open-source fine-tuning. \citet{lopez2023can} demonstrated that frontier models can forecast subsequent-day stock returns using raw headline text, while later approaches instruction-tune open source variants directly against financial tasks~\citep{koo2023finma}. However, these lines of work overwhelmingly evaluate the LLM as a standalone classifier (sentiment polarity, NER, QA).  By contrast, we treat the LLM as a frozen, zero-shot \emph{numerical} feature extractor---analogous to using a pre-trained vision model as a fixed encoder for downstream tasks---and evaluate its representations against a continuous downstream RL objective rather than NLP classification metrics.

\paragraph{Prompt \& Pipeline Optimization.}
Frameworks like DSPy~\citep{khattab2023dspy} formally compile language model calls by optimizing prompts against programmatic validation metrics. While these frameworks routinely optimize for exact-match accuracy or retrieval scores, our optimization loop extends this paradigm to a domain-specific continuous metric (rank correlation of predicted returns) prior to RL integration, treating the prompt as a discrete hyperparameter in the same spirit as architecture search.

\paragraph{Information Asymmetry and News Latency.}
Insider-trading research~\citep{seyhun1998investment} established that information edges decay rapidly as they become public.  For free-tier news feeds, institutional desks have already acted on the headline by the time a retail pipeline ingests it.  This latency shapes the horizon at which LLM-derived features can carry signal---a constraint we quantify in Section~\ref{sec:features}.

\section{System Architecture}
\label{sec:arch}

\begin{figure*}[t]
\centering
\begin{tikzpicture}[
  >=stealth,
  node distance=0.8cm and 0.6cm,
  font=\small\sffamily,
  box/.style={draw, rounded corners, align=center, thick, inner sep=6pt, minimum height=1.1cm},
  blueBox/.style={box, draw=blue!70!black, fill=blue!5},
  orangeBox/.style={box, draw=orange!80!black, fill=orange!5}
]
\node[blueBox] (sources) {News \& Flows\\RSS, EDGAR, Benzinga};
\node[blueBox, right=0.8cm of sources] (bundler) {Go Bundler\\(date $\times$ ticker)};
\node[blueBox, right=0.8cm of bundler] (llm) {LLM Feature Extractor\\(Qwen3-235B API)};
\draw[->, thick] (sources) -- (bundler);
\draw[->, thick] (bundler) -- (llm);
\node[blueBox, below=0.5cm of llm] (vector) {Feature Vector\\(9 Dims)};
\node[orangeBox, left=0.8cm of vector] (norm) {SB3 VecNormalize\\(clip\_obs=10)};
\node[orangeBox, left=0.8cm of norm] (ppo) {PPO Policy Agent\\(MLP Network)};
\node[orangeBox, left=0.8cm of ppo] (actions) {Trading Actions\\$\{-1, 0, +1\}$};
\draw[->, thick] (llm) -- (vector);
\draw[->, thick] (vector) -- (norm);
\draw[->, thick] (norm) -- (ppo);
\draw[->, thick] (ppo) -- (actions);
\end{tikzpicture}
\caption{System overview.  A Go-based ingestion pipeline collects news, filings, and macro data into a relational store.  The LLM produces per-ticker feature vectors, which the PPO agent consumes alongside OHLCV and technical indicators.}
\label{fig:arch}
\end{figure*}
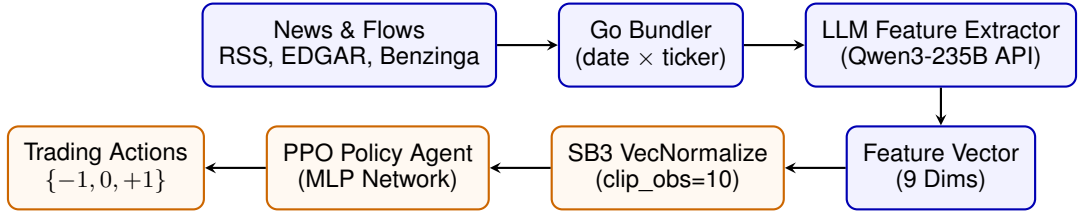

\subsection{Data Ingestion Pipeline}
Reproducible feature extraction requires a deterministic historical record.  We built a concurrent Go pipeline that ingests: (1)~news from Alpaca's Benzinga feed, (2)~RSS aggregations from financial outlets, and (3)~SEC EDGAR filings (Form~4 insider trades and 8-K disclosures).  Raw text is bundled per ticker per trading day and persisted to a SQLite database (the ``backfill layer'').

This backfill-first design prevents look-ahead bias: the LLM always reads from a frozen snapshot whose information boundary is strictly $\leq$~day~$d$.  It also allows the prompt-optimization loop (Section~\ref{sec:po2}) to re-extract features from identical text without re-scraping.

Because our pipeline enforces strict daily information boundaries (aggregating all feeds at the close of day $d$), it is intentionally blind to intra-day volatility and high-frequency market microstructure. Consequently, the RL agent operates on smoothed, day-over-day narrative shifts rather than instantaneous headline shocks. We measure the effect of this daily resolution in Section~\ref{sec:features}.

\subsection{Feature Schema}
\label{sec:schema}

A frontier LLM (\texttt{Qwen3 235B A22B Instruct 2507}) processes each ticker's daily bundle and outputs a structured JSON mapping into continuous RL observation bounds:
\begin{enumerate}
    \item \textbf{Stock-level LLM features (4 dims):}
    \begin{itemize}[nosep]
        \item \texttt{sentiment} $\in [-1, 1]$: Directional conviction from the daily news flow ($-1$: very bearish, $+1$: very bullish).
        \item \texttt{impact} $\in [0, 1]$: Financial materiality of the bundle (e.g., CEO resignation vs.\ routine marketing).
        \item \texttt{conflicting\_signals} $\in [0,1]$: Evidentiary contradiction across competing sources within the same bundle.
        \item \texttt{news\_novelty} $\in [0,1]$: Divergence of the current day's narrative from historical baselines.

    \end{itemize}
    \item \textbf{Macroeconomic features (5 dims):}
    \texttt{vix} (market anxiety), \texttt{treasury\_10y} (discount rate proxy), and \texttt{credit\_spread} (corporate default risk), sourced from FRED.  Two additional LLM-inferred regime flags (\texttt{market\_sentiment}, \texttt{macro\_event\_flag}) complement the systematic landscape.
\end{enumerate}

This JSON constraint grounds high-dimensional linguistic narratives into an explicit 9-dimensional vector digestible by the downstream MLP.  Table~\ref{tab:feature_distributions} details the summary statistics.

\input{tables/tab4_distributions.tex}

\subsection{RL Policy Agent}
We use FinRL~\citep{liu2020finrl} to construct the trading environment and PPO agent.  The composite state vector is:
\begin{equation}
S_t = \left[ P_t \;\Vert\; \mathcal{E}_t \;\Vert\; M_t \;\Vert\; B_t \right]
\end{equation}
where $P_t$ is OHLCV + technical indicators, $\mathcal{E}_t$ is ticker-level LLM features, $M_t$ is macro features, and $B_t$ is portfolio state.  Observations are normalized via \texttt{VecNormalize} ($\texttt{clip\_obs}=10$).

\begin{figure}[t]
\centering
\includegraphics[width=\columnwidth]{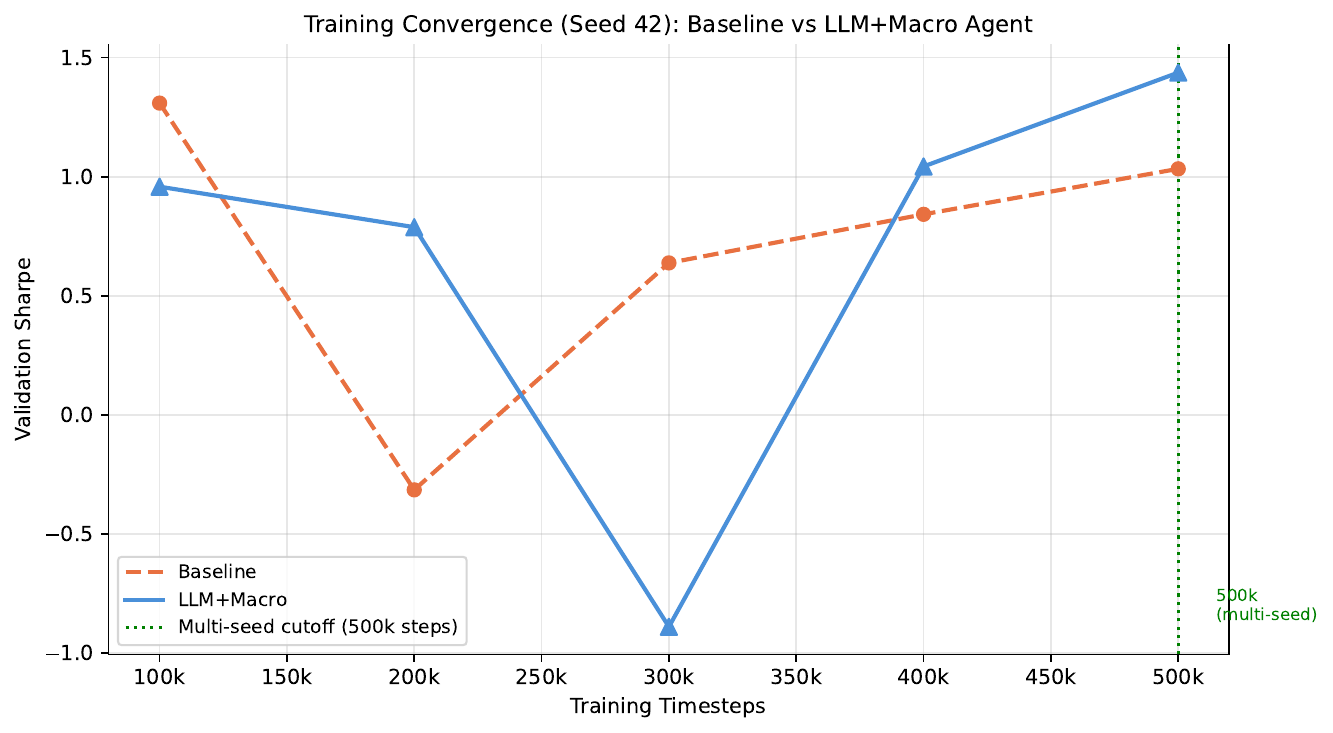}
\caption{Training convergence (seed 42).  Validation Sharpe at each 100k-step checkpoint on H1~2025.  Both agents plateau by 400--500k steps; the vertical line marks the multi-seed cutoff.}
\label{fig:convergence}
\end{figure}

The agent is a PPO with an MLP policy, trained for 500k timesteps on 2023--2024 data---a cutoff supported by the convergence analysis in Figure~\ref{fig:convergence}.  The RL framework trades a 21-ticker universe (${\sim}$10,500 rows).

All results are averaged across five seeds $\{0,1,2,3,42\}$ with deterministic hierarchical seeding (PyTorch \texttt{manual\_seed} + SB3 environment seeds) to ensure reproducibility.

\section{Prompt Optimization}
\label{sec:po2}

Standard prompt engineering does not scale when the downstream task is continuous-valued RL rather than classification.  NLP losses like BLEU or accuracy do not measure whether extracted features rank future returns correctly.  We designed an automated optimization loop that treats the extraction prompt as a discrete hyperparameter and tunes it against the Information Coefficient (IC).

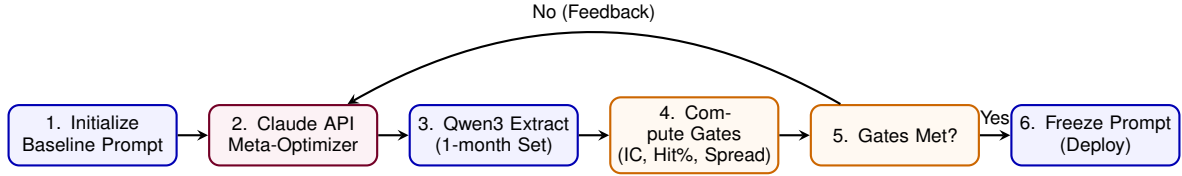
\begin{figure*}[t]
\centering
\begin{tikzpicture}[
  >=stealth,
  node distance=0.6cm and 0.3cm,
  font=\scriptsize\sffamily,
  box/.style={draw, rounded corners, align=center, thick, inner sep=3pt, minimum height=0.8cm, text width=2cm},
  procBox/.style={box, draw=blue!70!black, fill=blue!5},
  evalBox/.style={box, draw=orange!80!black, fill=orange!5},
  metaBox/.style={box, draw=purple!60!black, fill=purple!5}
]
\node[procBox] (baseline) {1. Initialize\\Baseline Prompt};
\node[metaBox, right=0.4cm of baseline] (claude) {2. Claude API\\Meta-Optimizer};
\node[procBox, right=0.4cm of claude] (extract) {3. Qwen3 Extract\\(1-month Set)};
\node[evalBox, right=0.4cm of extract] (eval) {4. Compute Gates\\(IC, Hit\%, Spread)};
\node[evalBox, right=0.4cm of eval] (check) {5. Gates Met?};
\node[procBox, right=0.4cm of check] (final) {6. Freeze Prompt\\(Deploy)};
\draw[->, thick] (baseline) -- (claude);
\draw[->, thick] (claude) -- (extract);
\draw[->, thick] (extract) -- (eval);
\draw[->, thick] (eval) -- (check);
\draw[->, thick] (check) -- node[above] {\scriptsize Yes} (final);
\draw[->, thick] (check) to[bend right=30] node[above] {\scriptsize No (Feedback)} (claude);
\end{tikzpicture}
\caption{Prompt optimization workflow.  Claude iteratively refines the extraction prompt based on downstream financial metric feedback.}
\label{fig:po2_flow}
\end{figure*}

The pipeline operates via a feedback loop with Anthropic's Claude API as meta-optimizer.  The workflow, illustrated in Figure~\ref{fig:po2_flow}, follows six steps:
\begin{enumerate}[nosep]
    \item \textbf{Initialize} a baseline chain-of-thought prompt (\texttt{v0}).
    \item \textbf{Define gates:} IC (rank correlation stability), Hit\% (directional accuracy), Quintile Spread (monotonicity of ranked portfolios).  These measure whether the LLM's numerical outputs predict future returns, unlike NLP metrics which only measure text quality.
    \item \textbf{Meta-optimize:} Claude suggests a discrete structural mutation (e.g., adding few-shot examples, redefining output ranges).
    \item \textbf{Extract:} Features for a one-month subset (January 2025, 769~bundles, 38~tickers) using the mutated prompt via Qwen3.
    \item \textbf{Evaluate:} Compute IC gates.
    \item \textbf{Iterate:} If gates fail, send metrics back to Claude as feedback; if passed, freeze the prompt.
\end{enumerate}

\input{tables/tab1_mutations.tex}

As shown in Table~\ref{tab:prompt_mutations}, we evaluated five distinct mutation hypotheses beyond the chain-of-thought baseline:
\begin{enumerate}[nosep, leftmargin=*]
    \item \textbf{mut1 (Impact-Surprise):} Redefines \texttt{impact} as the magnitude of market surprise relative to consensus.
    \item \textbf{mut2 (Few-Shot):} Adds three concrete calibration examples (priced-in beat, genuine surprise, and conflicting flow).
    \item \textbf{mut3 (Separate Reasoning):} Decouples reasoning into distinct news and flow signal analysis blocks.
    \item \textbf{mut4 (Counterfactual):} Uses a "what-if" counterfactual test to anchor sentiment scores against market defaults.
    \item \textbf{mut5 (Combined):} Merges mut1, mut2, and mut4 into a single prompt.
\end{enumerate}

The results demonstrate that explicit few-shot calibration (\texttt{mut2-few-shot}) was the critical driver of performance, improving IC~IR to $+0.104$ compared to the baseline's $-0.024$.  In contrast, combining multiple structural changes (\texttt{mut5}) caused a performance collapse, likely due to instruction bloat and conflicting reasoning anchors. The prompt was frozen as \texttt{v4-stable-core} (Figure~\ref{fig:prompt_template}) and used for all subsequent extraction.

\subsection{Adequacy Gates}
Before committing to RL training, we evaluated the extracted features against four predictive adequacy gates (Table~\ref{tab:adequacy_gates}). The prompt cleared all four thresholds, notably exceeding the IC requirement by a substantial margin. Because our optimization loop explicitly targets downstream predictive utility, this strong rank correlation confirms the prompt's validity in generating a continuous observation space for the downstream RL agent.

\input{tables/tab2_adequacy_gate.tex}

\begin{figure*}[t]
\centering
\begin{tcolorbox}[colback=blue!3!white, colframe=blue!40!black, title=Prompt \texttt{v1-stable-core} Feature Extraction Template, boxrule=0.8pt]
\footnotesize
\textbf{System:} You are a quantitative feature extraction engine for an institutional trading system. Given an event bundle about a specific ticker, output ONLY valid JSON with numerical features...

\textbf{User:} Extract trading features for ticker \{\{.Ticker\}\} from the following event bundle on \{\{.Date\}\}. Consider ALL signal types together --- news articles, insider trades, and options flow. Output ONLY valid JSON matching this schema.

\textbf{Field definitions} (fill in this order):
\begin{itemize}[nosep, leftmargin=*]
\item \texttt{reasoning}: Write ONE sentence summarizing the key signal and WHY it moves the stock.
\item \texttt{sentiment}: Predicted FUTURE PRICE TRAJECTORY over the next 1-5 trading days. [-1.0 strongly bearish, 0.0 no edge, +1.0 strongly bullish] Focus on SURPRISE vs CONSENSUS, not absolute tone.
\item \texttt{impact}: Materiality of the news. [0.0 trivial, 1.0 highly market-moving].
\item \texttt{conflicting\_signals}: Do the signals point in contradictory directions? [0.0 aligned, 1.0 strongly contradictory].
\item \texttt{insider\_trading}: Insider BUYS are a moderately bullish signal. Insider SELLS are WEAK (usually scheduled 10b5-1 plans).
\end{itemize}

\textbf{Calibration Example A --- Priced-in beat:}
\textit{``AAPL reports Q4 earnings beating consensus by 2\%, in line with whisper numbers.''}
$\rightarrow$ \texttt{sentiment: +0.1, impact: 0.2, conflicting\_signals: 0.0}

\textbf{Calibration Example B --- Genuine surprise:}
\textit{``NVDA unexpectedly raises full-year guidance 40\% above Street estimates...''}
$\rightarrow$ \texttt{sentiment: +0.8, impact: 0.9, conflicting\_signals: 0.1}
\end{tcolorbox}
\caption{Abbreviated visualization of \texttt{v1-stable-core}.  Few-shot calibration examples anchoring numerical output ranges proved necessary to maximize IC.}
\label{fig:prompt_template}
\end{figure*}

\section{Empirical Evaluation}
\label{sec:results}

Training spans 2023--2024.  Validation covers 2025\,H1 (January--June, 120~trading days), a period of tariff-driven macro volatility.  The held-out test covers 2025\,H2 (July--December), a materially calmer regime.

\subsection{Feature Validity}
\label{sec:features}

Before examining downstream RL performance, we verify that the optimized prompt produces features with genuine predictive signal.  Table~\ref{tab:feature_ic} reports the Information Coefficient (IC)---the Spearman rank correlation between each feature's daily values and subsequent 5-day returns, averaged across trading days.  IC normalizes by the standard deviation of daily ICs, analogous to a signal-to-noise ratio.  Among the LLM-derived features, \texttt{conflicting\_signals} (IC~$=0.233$, $t=2.52$) and \texttt{impact} ($0.177$, $t=1.91$) carry the strongest signal.

\input{tables/tab3_feature_ic.tex}

\begin{figure*}[t]
  \centering
  \begin{minipage}{0.48\textwidth}
    \centering
    \includegraphics[width=\linewidth]{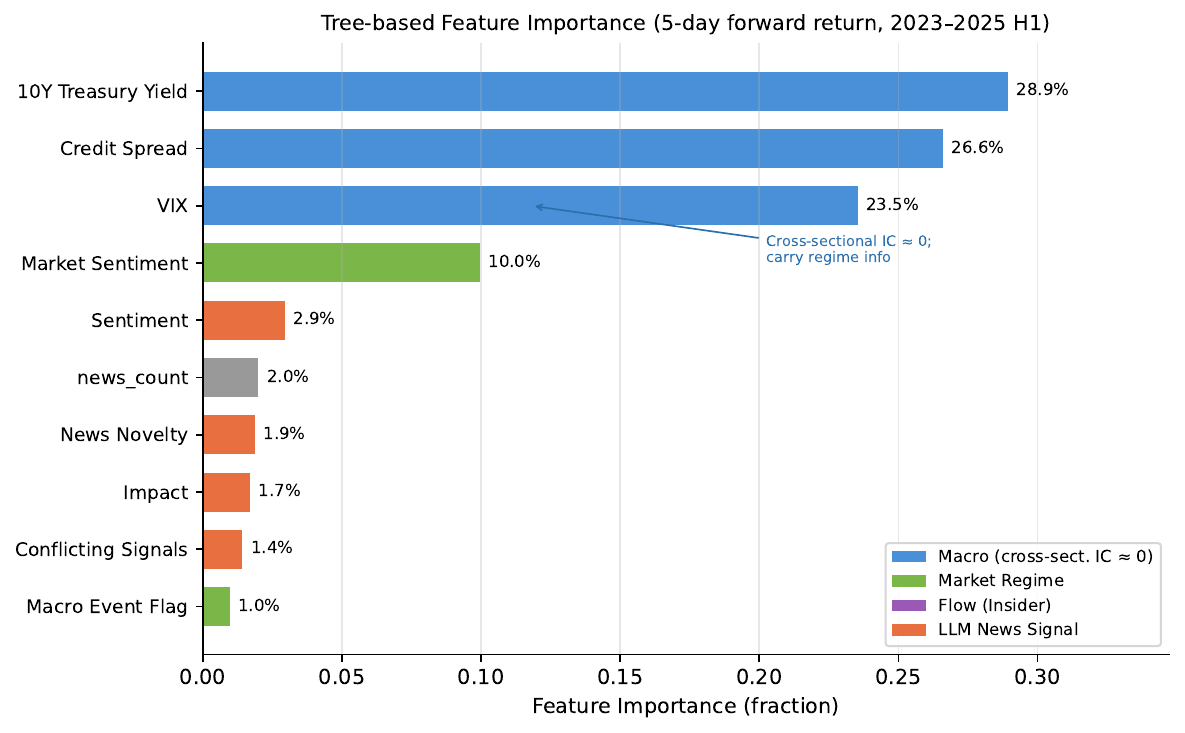}
    \caption{Feature importance from a gradient-boosted tree on 5-day forward returns. Macro features dominate (credit\_spread 28\%, VIX 16\%, treasury 14\%) despite zero cross-sectional IC.}
    \label{fig:importance}
  \end{minipage}\hfill
  \begin{minipage}{0.48\textwidth}
    \centering
    \includegraphics[width=\linewidth]{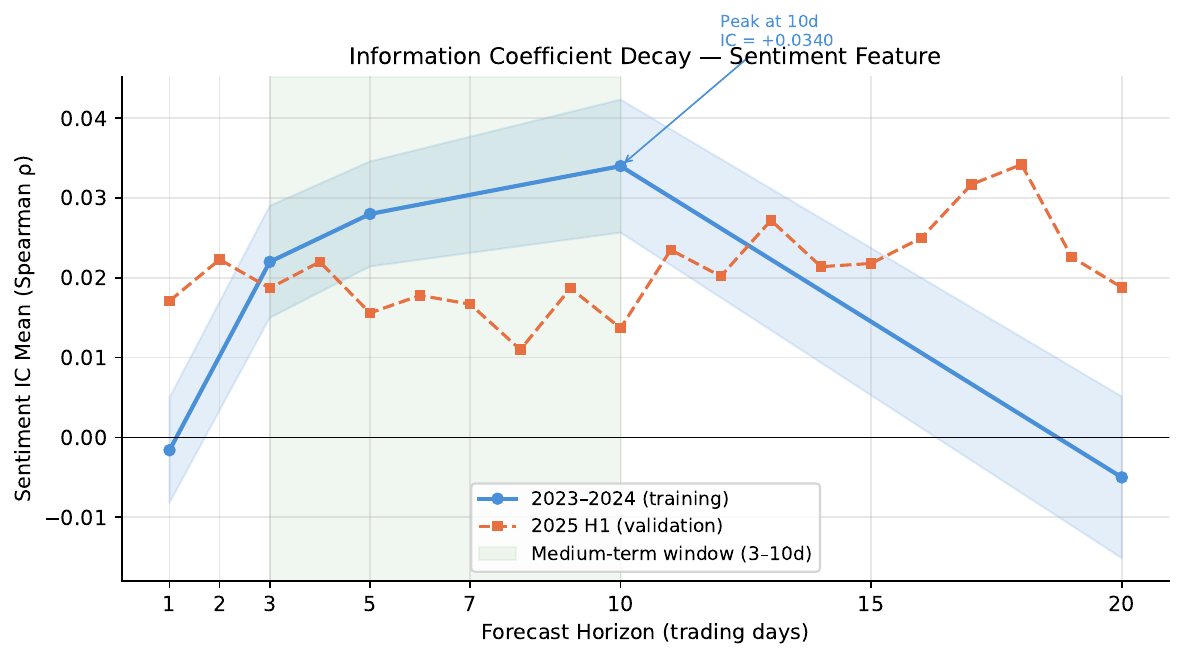}
    \caption{IC decay for sentiment across horizons. Signal peaks at 3--10\,days, consistent with the delayed-news constraint.}
    \label{fig:ic_decay}
  \end{minipage}
\end{figure*}

The macro features (VIX, Treasury, credit spread) register IC~$\approx 0$ by construction: they are identical across all tickers on a given day, so cross-sectional rank correlation is undefined.  Their value is purely time-serial---they tell the RL agent \emph{when} to trade cautiously, not \emph{which} ticker to favor.  Figure~\ref{fig:importance} confirms this asymmetry: a gradient-boosted tree trained on 5-day forward returns assigns 58\% cumulative importance to macro features despite their zero cross-sectional IC.

Figure~\ref{fig:ic_decay} shows sentiment IC as a function of forecast horizon. The signal is near zero at one day, peaks at 3--10\,days, and decays by day~20---consistent with the delayed-news constraint: the 1-day edge has been captured by faster participants, leaving only medium-term narrative drift.

\subsection{Validation: Macro-Shock Regime (H1 2025)}
\label{sec:h1_results}

Cross-sectional feature evaluation is conducted on a broader 38-ticker US large-cap signal universe, while the downstream RL environment restricts execution to a 21-ticker liquid trading subset plus SPY; the exact ticker lists are provided in Appendix~\ref{app:universes}.

Given that the LLM features carry genuine signal, we now ask whether this translates into downstream RL performance.  Table~\ref{tab:ablation} presents the four-configuration ablation on H1~2025.  No configuration significantly outperforms the price-only baseline (all paired $t$-test $p>0.1$).

\input{tables/tab6_ablation}
\input{tables/tab8_test_results.tex}

\begin{figure}[t]
\centering
\includegraphics[width=\columnwidth]{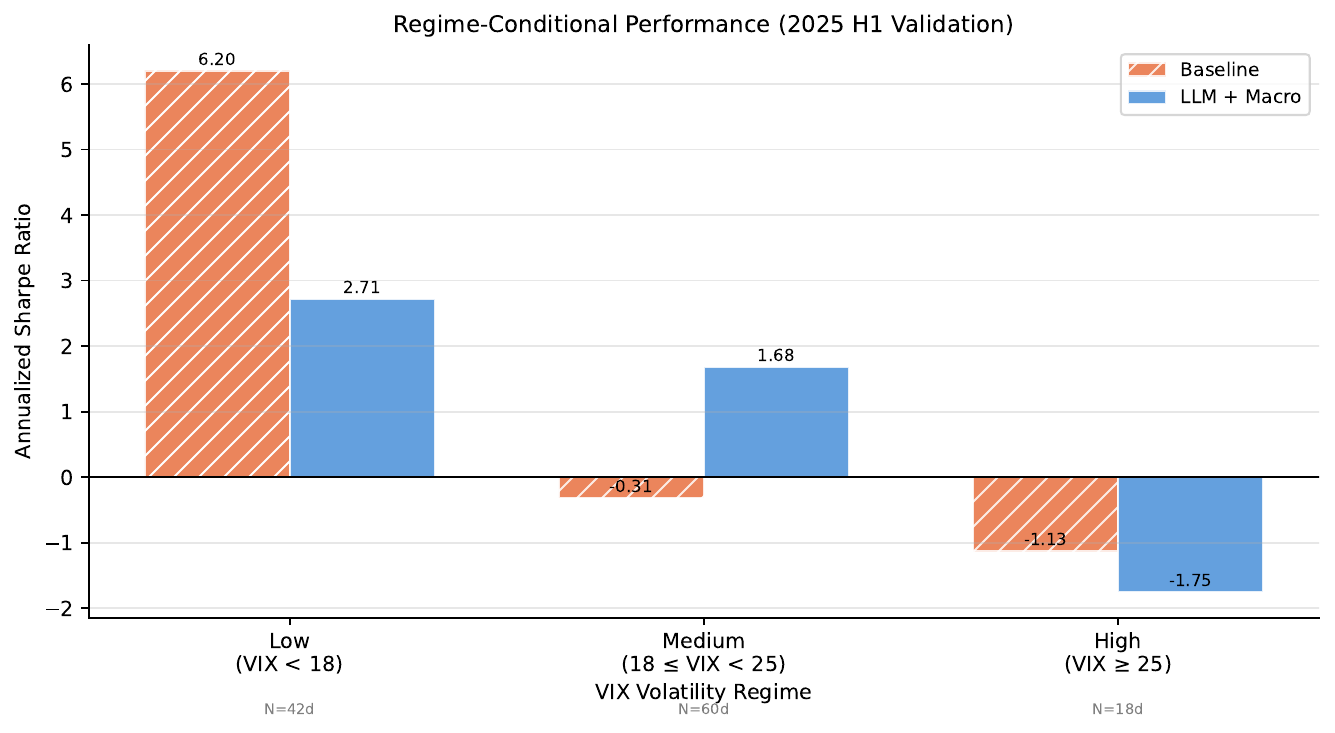}
\caption{Regime-conditional performance.  During elevated VIX (shaded), the LLM-augmented agent under-performs the baseline.  In low-volatility windows the gap narrows or reverses.}
\label{fig:regime}
\end{figure}

The pattern is informative.  LLM-only is the worst configuration (Sharpe~$-0.411$): trading on idiosyncratic stock news during a systemic shock amounts to ignoring the dominant risk factor.  Macro-only ($-0.007$) tracks the baseline closely.  LLM+Macro ($-0.267$) is better than LLM-only because the macro features provide a ``regime brake,'' but the noisy LLM signals still drag it below baseline.  Figure~\ref{fig:regime} visualizes this: during elevated VIX, the LLM-augmented agent systematically under-performs.

\subsection{Held-Out Test: Calm Regime (H2 2025)}
\label{sec:h2_results}

If idiosyncratic news fails only when macro risk dominates, a calmer regime should restore its value.  We locked all model parameters and evaluated once on H2~2025 (Table~\ref{tab:test_results}).

All three augmented configurations now exceed the baseline (Sharpe~0.809).  LLM-only recovers to 1.001, suggesting that stock-level narratives can capture signal when systemic risk subsides.  However, Macro-only remains strongest (1.099, $\Delta=+0.290$), and LLM+Macro ($1.038$, $p=0.49$ vs.\ baseline) does not reach statistical significance at $N=5$ seeds.  SPY buy-and-hold (1.756) outperforms all RL configurations during the late-2025 bull rally; our contribution is the \emph{relative} ablation, not absolute excess return.

\subsection{Robustness Checks}

\paragraph{Transaction cost sensitivity.}
Table~\ref{tab:cost_sensitivity} varies transaction costs from 0 to 50\,bp for seed~42 on H1~2025.  The baseline's advantage over LLM+Macro is stable across all cost levels, confirming that the H1 null result is not an artifact of unrealistic friction assumptions.

\input{tables/tab7_cost_sensitivity.tex}

\section{Discussion}
\label{sec:discussion}

\paragraph{The feature-to-policy gap.}
Our results expose a gap between intermediate representation quality and downstream task performance---a phenomenon familiar from transfer learning, where strong pre-trained features do not guarantee fine-tuning success under distribution shift.  The prompt optimization loop successfully produces features with high rank correlation to future returns.  However, this signal only translates into RL policy improvement when the test distribution resembles training conditions.  When a macroeconomic shock shifts the data distribution (H1~2025), the RL agent cannot exploit features whose predictive structure has changed.

\paragraph{Macro features as a regime brake.}
VIX, Treasury yields, and credit spreads do not predict \emph{which} stocks will outperform---their cross-sectional IC is zero (Table~\ref{tab:feature_ic}).  Instead, they tell the RL agent when the market environment has shifted, allowing it to reduce exposure.  Feature importance analysis (Figure~\ref{fig:importance}) confirms that the agent assigns 58\% of its decision weight to macro inputs.  This time-serial conditioning is invisible to IC-based evaluation but critical to policy performance.

\paragraph{Information latency as a feature constraint.}
Our reliance on free-tier news feeds bounds the temporal resolution of extractable signal.  The IC-decay curve (Figure~\ref{fig:ic_decay}) shows that predictive power peaks at 3--10\,days.  From an ML perspective, this is an input-quality constraint analogous to training on low-resolution images---the representation is valid but resolution-limited.

\paragraph{Limitations.}
\begin{itemize}[nosep]
  \item \textbf{Statistical power.}  Five seeds on a 120-day window yield low power ($<$50\% to detect $\Delta$Sharpe$=$0.3 at $\sigma=0.4$).
  \item \textbf{Absolute performance.}  All agents under-perform SPY buy-and-hold.  The contribution is the relative ablation, not absolute return.
  \item \textbf{Narrow universe.}  21~large-cap US equities.  Generalization is an open question.
  \item \textbf{Input resolution.}  A low-latency news feed may restore short-horizon signal.
\end{itemize}

\section{Conclusion}
\label{sec:conclusion}

We present an automated prompt-optimization pipeline that successfully tunes a frozen LLM to produce predictive numerical features for an RL trading agent.  However, valid intermediate representations do not automatically yield downstream task improvement---a finding consistent with known challenges in transfer learning under distribution shift.  The gap between feature quality and policy robustness is regime-dependent: LLM-derived features help when the test distribution is stable but degrade when macroeconomic shocks shift the data-generating process.

More broadly, our results suggest that evaluating LLM-generated features on intermediate metrics alone is insufficient.  Multi-regime out-of-sample testing, with explicit distribution-shift stress periods, should become a standard protocol for any pipeline that feeds LLM representations into a downstream learner.

\appendix
\section{Ticker Universes}
\label{app:universes}

\paragraph{Signal universe (38 tickers).}
LLM feature extraction and prompt-optimization experiments use the following 38-ticker signal universe:
\begin{quote}
\footnotesize
AAPL, ABBV, ADBE, AMD, AMZN, AVGO, BA, BAC, CAT, COST, CRM, CVX, GE, GOOGL, GS, HD, INTC, IWM, JNJ, JPM, LLY, MA, MCD, META, MSFT, NFLX, NKE, NVDA, ORCL, QCOM, QQQ, RTX, SPY, TSLA, UNH, V, WMT, XOM.
\end{quote}

\paragraph{Tradable RL universe (21 tickers).}
The downstream RL environment restricts execution to the following 21-ticker liquid trading universe:
\begin{quote}
\footnotesize
AAPL, MSFT, AMZN, NVDA, META, TSLA, AMD, NFLX, ADBE, QCOM, JPM, V, MA, GS, UNH, LLY, XOM, WMT, BA, CAT, SPY.
\end{quote}

\paragraph{Why the universes differ.}
The broader 38-ticker universe improves cross-sectional feature evaluation and includes contextual instruments such as QQQ, IWM, and GOOGL.  The RL agent trades only the 21-ticker liquid subset.  GOOGL is excluded from trading for compliance reasons, although its signals remain available in the upstream feature store.

\FloatBarrier
\bibliography{references}

\end{document}

%% file: tables/tab4_distributions.tex
\begin{table*}[t]
\centering
\caption{Feature Value Distributions (2023-2024 training period)}
\label{tab:feature_distributions}
\begin{tabular}{@{}llcccccc@{}}
\toprule
Feature & Group & Mean & Std & Min & Max & \% Non-zero & Scale \\
\midrule
sentiment & LLM News & -0.007 & 0.290 & -1.000 & 0.900 & 71.8\% & [-1, 1] \\
impact & LLM News & 0.292 & 0.233 & 0.000 & 1.000 & 71.9\% & [0, 1] \\
\texttt{conflicting\_signals} & LLM News & 0.076 & 0.182 & 0.000 & 1.000 & 18.5\% & [0, 1] \\
\texttt{news\_novelty} & LLM News & 0.737 & 0.424 & 0.000 & 1.000 & 77.9\% & [0, 1] \\
\midrule
\texttt{market\_sentiment} & Market Regime & 0.175 & 0.405 & -1.000 & 0.800 & 98.7\% & [-1, 1] \\
\texttt{macro\_event\_flag} & Market Regime & 0.481 & 0.500 & 0.000 & 1.000 & 48.1\% & \{0, 1\} \\
\midrule
\texttt{treasury\_10y} & Macro & 4.085 & 0.370 & 3.300 & 4.980 & 100.0\% & [3.5, 5.5] \\
vix & Macro & 16.120 & 3.262 & 11.860 & 38.570 & 100.0\% & [12, 80] \\
\texttt{credit\_spread} & Macro & 3.675 & 0.620 & 2.600 & 5.220 & 100.0\% & [0.5, 3.0] \\
\midrule
\multicolumn{8}{@{}p{\linewidth}@{}}{\textit{VecNormalize (clip\_obs=10) applied during training normalizes these distributions.}} \\
\bottomrule
\end{tabular}
\end{table*}

%% file: tables/tab1_mutations.tex
\begin{table*}[t]
  \centering
  \caption{Prompt mutation results.  Top: January 2025 optimization set (769 bundles, 38 tickers).  Bottom: February 2025 out-of-sample validation.}
  \label{tab:prompt_mutations}

  \begin{subtable}{\textwidth}
    \centering
    \caption{Jan 2025 --- Optimization Set}
    \begin{tabular}{llccccc}
    \toprule
    Candidate & Hypothesis & IC IR & Hit\% & Spread & Brier & Comp. \\
    \midrule
    v3-baseline & Chain-of-thought & $-$0.024 & 45.5 & $-$1.07\% & 0.305 & $-$0.144 \\
    mut1 & Impact = surprise & $-$0.075 & 22.2 & $-$1.66\% & 0.291 & $-$0.325 \\
    \rowcolor{yellow!20}
    mut2-few-shot$^\dagger$ & Few-shot examples & \textbf{+0.104} & 71.4 & +0.22\% & 0.288 & +0.191 \\
    mut3-separate & Separate reasoning & +0.029 & 60.0 & +0.75\% & 0.300 & +0.134 \\
    mut4 & Counterfactual & $-$0.098 & 57.1 & $-$0.29\% & 0.289 & +0.052 \\
    mut5-combined & mut1+2+4 & $-$0.175 & 55.6 & $-$2.06\% & 0.290 & $-$0.243 \\
    \bottomrule
    \end{tabular}
    
    \vspace{4pt}
    {\footnotesize $^\dagger$Frozen as \texttt{v4-stable-core}.}
  \end{subtable}

  \vspace{1.5em}
  
  \begin{subtable}{\textwidth}
    \centering
    \caption{Feb 2025 --- OOS Validation}
    \begin{tabular}{lccccc}
    \toprule
    Candidate & IC IR & Hit\% & Spread & Brier & Comp. \\
    \midrule
    v3-baseline & $-$0.212 & 50.0 & N/A & 0.293 & $-$0.210 \\
    \rowcolor{yellow!20}
    mut2-few-shot & $-$0.044 & 25.0 & N/A & 0.281 & $-$0.230 \\
    \bottomrule
    \end{tabular}
  \end{subtable}
\end{table*}

%% file: tables/tab2_adequacy_gate.tex
\begin{table}[ht]
  \centering
  \caption{Adequacy gate assessment (\texttt{mut2-few-shot}, Jan 2025).}
  \label{tab:adequacy_gates}
  \resizebox{\columnwidth}{!}{%
  \begin{tabular}{@{}lccc@{}}
  \toprule
  Metric & Gate Threshold & Value & Status \\
  \midrule
  \cellcolor{green!15} signal\_coverage & $\geq 0.25$ & 0.408 & \textbf{PASS} \\
  \cellcolor{green!15} ic\_ir\_5d & $\geq 0.05$ & +0.104 & \textbf{PASS} \\
  \cellcolor{green!15} quintile\_spread & $> 0$ & +0.002 & \textbf{PASS} \\
  \cellcolor{green!15} hit\_rate & $\geq 0.52$ & 0.714 & \textbf{PASS} \\
  \bottomrule
  \end{tabular}%
  }

  \vspace{4pt}
  {\footnotesize $^\dagger$Structural issue: \texttt{pred\_prob} uses impact as confidence, but impact measures materiality, not prediction certainty.}
\end{table}

%% file: tables/tab3_feature_ic.tex
\begin{table*}[t]
  \centering
  \caption{Feature IC analysis (5-day forward return, 2023--2024 training period).  Cross-sectional IC for macro features is zero by construction: they are constant across tickers on any given day and carry regime information detectable only by the RL policy, not by cross-sectional ranking.}
  \label{tab:feature_ic}
  \begin{tabular}{llrrrrc}
  \toprule
  Feature & Group & IC Mean & IC IR & $t$-stat & \% Pos & $N$ \\
  \midrule
  sentiment & LLM News & 0.016 & \textbf{0.093} & 1.00 & 0.5 & 117 \\
  impact & LLM News & 0.029 & \textbf{0.177} & 1.91 & 0.6 & 117 \\
  conflicting\_signals & LLM News & 0.040 & \textbf{0.233} & 2.52 & 0.6 & 117 \\
  news\_novelty & LLM News & 0.011 & \textbf{0.065} & 0.70 & 0.5 & 117 \\
  \bottomrule
  \end{tabular}
\end{table*}

%% file: tables/tab6_ablation.tex
\begin{table*}[t]
  \centering
  \caption{H1 2025 validation ablation ($N\!=\!5$ seeds, 120 trading days). Mean$\pm$std across seeds. $\Delta$Sharpe vs.\ Baseline.}
  \label{tab:ablation}
  \begin{tabular}{lcrrrr}
  \toprule
  Config & N Feats & Sharpe & Return\% & Max DD\% & $\Delta$Sharpe \\
  \midrule
  Baseline (OHLCV + Tech) & 0 & 0.010 $\pm$ 0.618 & -5.03 $\pm$ 18.00 & 34.91 $\pm$ 6.01 & — \\
  LLM Signals Only$^\dag$ & 6 & -0.411 $\pm$ 0.690 & -16.42 $\pm$ 18.81 & 42.12 $\pm$ 8.11 & -0.421 \\
  \rowcolor{blue!8}Macro Only$^\ddag$ & 5 & -0.007 $\pm$ 0.355 & -6.16 $\pm$ 9.74 & 34.78 $\pm$ 9.28 & -0.017 \\
  LLM + Macro (Full) & 10 & -0.267 $\pm$ 0.284 & -13.91 $\pm$ 7.41 & 38.66 $\pm$ 6.34 & -0.276 \\
  \bottomrule
  \end{tabular}
  \vspace{2pt}
  \par\footnotesize
    $^\dag$~LLM-only: sentiment, impact, conflicting\_signals, news\_novelty + 2 regime flags.
    $^\ddag$~Macro-only: VIX, treasury\_10y, credit\_spread + 2 regime flags.
    No config beats Baseline ($p>0.1$, paired $t$, $N\!=\!5$).
\end{table*}

%% file: tables/tab8_test_results.tex
\begin{table*}[t]
  \centering
  \caption{H2 2025 held-out test ($N\!=\!5$ seeds). Mean$\pm$std. $\Delta$Sharpe vs.\ Baseline.}
  \label{tab:test_results}
  \begin{tabular}{lcrrrr}
  \toprule
  Config & N Feats & Sharpe & Return\% & Max DD\% & $\Delta$Sharpe \\
  \midrule
  Baseline (OHLCV + Tech) & 0 & 0.809 $\pm$ 0.333 & 11.29 $\pm$ 6.21 & 14.96 $\pm$ 4.04 & — \\
  LLM Signals Only$^\dag$ & 5 & 1.001 $\pm$ 0.853 & 12.98 $\pm$ 8.93 & 16.27 $\pm$ 6.59 & +0.192 \\
  Macro Only$^\ddag$ & 5 & 1.099 $\pm$ 0.695 & 16.04 $\pm$ 9.53 & 15.30 $\pm$ 3.15 & +0.290 \\
  \rowcolor{yellow!20}LLM + Macro (Full) & 10 & 1.038 $\pm$ 0.424 & 15.14 $\pm$ 6.19 & 17.27 $\pm$ 2.18 & +0.229 \\
  SPY buy-and-hold$^\S$ & — & 1.756$^\S$ & 11.87$^\S$ & 5.07$^\S$ & +0.947 \\
  \midrule
  \multicolumn{6}{l}{Paired $t$-test (LLM+Macro vs. Baseline): $t(4)=0.76$, $p=0.4873$} \\
  \bottomrule
  \end{tabular}
  \vspace{2pt}
  \par\footnotesize
    $^\dag$/$^\ddag$~Feature groups as in Table~\ref{tab:ablation}.
    $^\S$~SPY buy-and-hold (single value).
    Models fixed before unlocking test set.
\end{table*}

%% file: tables/tab7_cost_sensitivity.tex
\begin{table*}[t]
\centering
\caption{Transaction cost sensitivity (2025 H1, seed 42). The baseline advantage over LLM+Macro is robust across all cost levels, confirming the H1 null result is not a friction artifact.}
\label{tab:cost_sensitivity}
\begin{tabular}{rrcccc}
\toprule
Cost (bp) & Cost (\%) & Baseline Sharpe & LLM+Macro Sharpe & $\Delta$ Sharpe & LLM Win? \\
\midrule
0 & 0.00\% & 1.231 & 0.753 & -0.478 & \cellcolor{red!20}\texttimes \\
5 & 0.05\% & 1.202 & 0.730 & -0.472 & \cellcolor{red!20}\texttimes \\
\rowcolor{yellow!15}
10 & 0.10\% & 1.156 & 0.674 & -0.482 & \cellcolor{red!20}\texttimes \\
20 & 0.20\% & 1.108 & 0.549 & -0.559 & \cellcolor{red!20}\texttimes \\
50 & 0.50\% & 0.929 & 0.580 & -0.349 & \cellcolor{red!20}\texttimes \\
\bottomrule
\end{tabular}
\par\vspace{1ex}
\footnotesize
\textit{$\Delta$ Sharpe = LLM Sharpe - Baseline Sharpe. Standard cost (0.10\%) highlighted.}
\end{table*}